\documentclass[10pt,twocolumn,letterpaper]{article}

\usepackage{cvpr}
\usepackage{times}
\usepackage{epsfig}
\usepackage{graphicx}
\usepackage{amsmath}
\usepackage{amssymb}
\usepackage{comment}

\usepackage[pagebackref=true,breaklinks=true,letterpaper=true,colorlinks,bookmarks=false]{hyperref}

\cvprfinalcopy 

\def\cvprPaperID{741} 

\newcommand{\larry}[1]{\textcolor{blue}{#1 --Larry}}
\newcommand{\xinlei}[1]{\textcolor{red}{#1 --Xinlei}}
\newcommand{\vtf}{\mathbf{\tilde{v}}}
\newcommand{\vf}{\mathbf{v}}
\newcommand{\uf}{\mathbf{u}}
\newcommand{\wf}{\mathbf{w}}
\newcommand{\ssf}{\mathbf{s}}
\newcommand{\wtf}{\mathbf{\tilde{w}}}
\ifcvprfinal\fi
\begin{document}

\title{Learning a Recurrent Visual Representation for Image Caption Generation}

\author{Xinlei Chen\\
Carnegie Mellon University\\
{\tt\small xinleic@cs.cmu.edu}
\and
C. Lawrence Zitnick\\
Microsoft Research, Redmond\\
{\tt\small larryz@microsoft.com}
}

\maketitle

\begin{abstract}
In this paper we explore the bi-directional mapping between images and their sentence-based descriptions. We propose learning this mapping using a recurrent neural network. Unlike previous approaches that map both sentences and images to a common embedding, we enable the generation of novel sentences given an image. Using the same model, we can also reconstruct the visual features associated with an image given its visual description. We use a novel recurrent visual memory that automatically learns to remember long-term visual concepts to aid in both sentence generation and visual feature reconstruction. We evaluate our approach on several tasks. These include sentence generation, sentence retrieval and image retrieval. State-of-the-art results are shown for the task of generating novel image descriptions. When compared to human generated captions, our automatically generated captions are preferred by humans over $19.8\%$ of the time. Results are better than or comparable to state-of-the-art results on the image and sentence retrieval tasks for methods using similar visual features.

\end{abstract}

\section{Introduction}
A good image description is often said to ``paint a picture in your mind's eye.'' The creation of a mental image may play a significant role in sentence comprehension in humans \cite{just2004imagery}. In fact, it is often this mental image that is remembered long after the exact sentence is forgotten \cite{paivio1968pictures,lieberman1965words}. What role should visual memory play in computer vision algorithms that comprehend and generate image descriptions?

Recently, several papers have explored learning joint feature spaces for images and their descriptions \cite{hodosh2013framing,socher2013grounded,karpathy2014deep}. These approaches project image features and sentence features into a common space, which may be used for image search or for ranking image captions. Various approaches were used to learn the projection, including Kernel Canonical Correlation Analysis (KCCA) \cite{hodosh2013framing}, recursive neural networks \cite{socher2013grounded}, or deep neural networks \cite{karpathy2014deep}. While these approaches project both semantics and visual features to a common embedding, they are not able to perform the inverse projection. That is, they cannot generate novel sentences or visual depictions from the embedding.

In this paper, we propose a bi-directional representation capable of generating both novel descriptions from images and visual representations from descriptions. Critical to both of these tasks is a novel representation that dynamically captures the visual aspects of the scene that have already been described. That is, as a word is generated or read the visual representation is updated to reflect the new information contained in the word. We accomplish this using Recurrent Neural Networks (RNNs) \cite{elman1990finding,mikolov2010recurrent,mikolov2012context}. One long-standing problem of RNNs is their weakness in remembering concepts after a few iterations of recurrence. For instance RNN language models often find difficultly in learning long distance relations \cite{bengio1994learning,mikolov2010recurrent} without specialized gating units \cite{hochreiter1997long}. During sentence generation, our novel dynamically updated visual representation acts as a long-term memory of the concepts that have already been mentioned. This allows the network to automatically pick salient concepts to convey that have yet to be spoken. As we demonstrate, the same representation may be used to create a visual representation of a written description.

We demonstrate our method on numerous datasets. These include the PASCAL sentence dataset \cite{Rashtchian2010}, Flickr 8K \cite{Rashtchian2010}, Flickr 30K \cite{Rashtchian2010}, and the Microsoft COCO dataset \cite{lin2014microsoft}. When generating novel image descriptions, we demonstrate state-of-the-art results as measured by both BLEU \cite{papineni2002bleu} and METEOR \cite{banerjee2005meteor} on PASCAL 1K. Surprisingly, we achieve performance only slightly below humans as measured by BLEU and METEOR on the MS COCO dataset. Qualitative results are shown for the generation of novel image captions. We also evaluate the bi-directional ability of our algorithm on both the image and sentence retrieval tasks. Since this does not require the ability to generate novel sentences, numerous previous papers have evaluated on this task. We show results that are better or comparable to previous state-of-the-art results using similar visual features.

\begin{figure*}
  \centering
  \includegraphics[width=\linewidth]{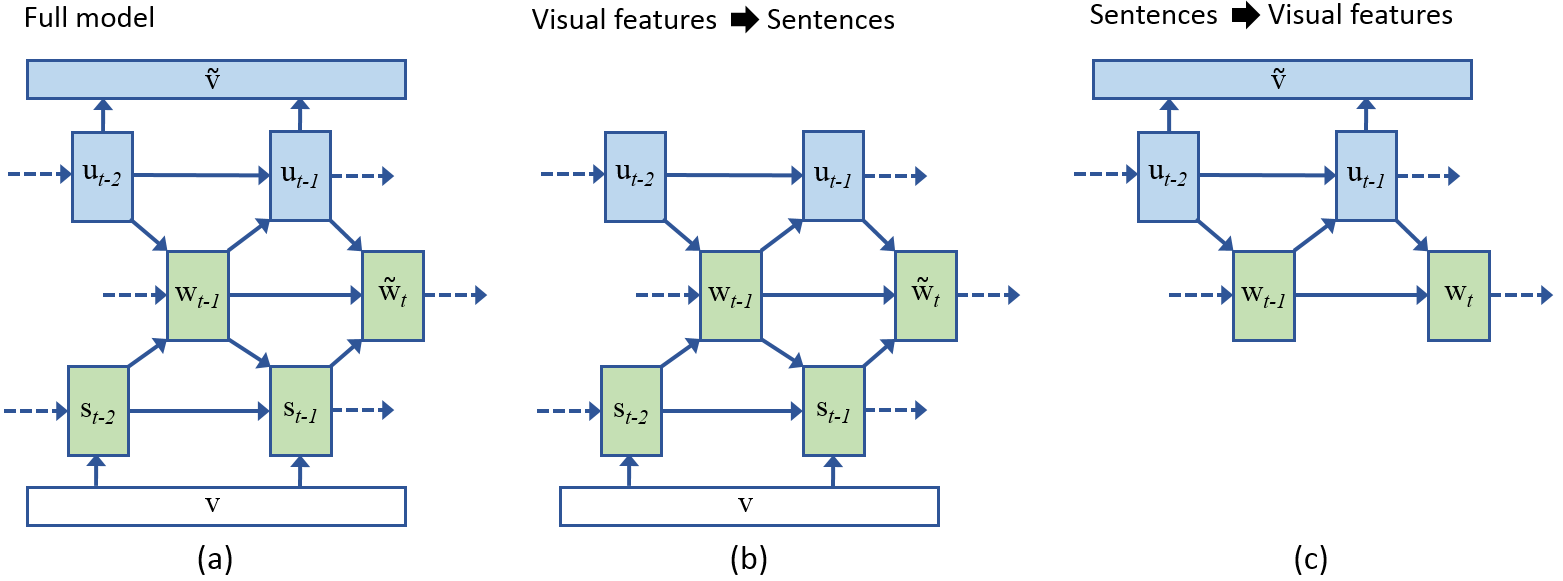}\\
  \caption{Illustration of our model. (a) shows the full model used for training. (b) and (c) show the parts of the model needed for generating sentences from visual features and generating visual features from sentences respectively.}\label{fig:model}
\end{figure*}

\section{Related work}

The task of building a visual memory lies at the heart of two long-standing AI-hard problems: grounding natural language symbols to the physical world and semantically understanding the content of an image. 
Whereas learning the mapping between image patches and single text labels remains a popular topic in computer vision~\cite{krizhevsky2012imagenet,frome2013devise,girshickCVPR14}, there is a growing interest in using entire sentence descriptions together with pixels to learn joint embeddings~\cite{hodosh2013framing,socher2013grounded,karpathy2014deep,gong2014improving}. Viewing corresponding text and images as correlated, KCCA~\cite{hodosh2013framing} is a natural option to discover the shared features spaces. However, given the highly non-linear mapping between the two, finding a generic distance metric based on shallow representations can be extremely difficult. Recent papers seek better objective functions that directly optimize the ranking~\cite{hodosh2013framing}, or directly adopts pre-trained representations~\cite{socher2013grounded} to simplify the learning, or a combination of the two~\cite{karpathy2014deep,gong2014improving}.

With a good distance metric, it is possible to perform tasks like bi-directional image-sentence retrieval. However, in many scenarios it is also desired to generate novel image descriptions and to hallucinate a scene given a sentence description. Numerous papers have explored the area of generating novel image descriptions \cite{farhadi2010every,yang2011corpus,kulkarni2011baby,yao2010i2t,mitchell2012midge,gupta2012choosing,kuznetsova2012collective,ryan2014multimodal}.
These papers use various approaches to generate text, such as using pre-trained object detectors with template-based sentence generation \cite{yang2011corpus,farhadi2010every,kulkarni2011baby}. Retrieved sentences may be combined to form novel descriptions \cite{kuznetsova2012collective}. Recently, purely statistical models have been used to generate sentences based on sampling \cite{ryan2014multimodal} or recurrent neural networks \cite{mao2014explain}. While \cite{mao2014explain} also uses a RNN, their model is significantly different from our model. Specifically their RNN does not attempt to reconstruct the visual features, and is more similar to the contextual RNN of \cite{mikolov2012context}. For the synthesizing of images from sentences, the recent paper by Zitnick \etal \cite{zitnick2013bringing} uses abstract clip art images to learn the visual interpretation of sentences. Relation tuples are extracted from the sentences and a conditional random field is used to model the visual scene.

There are numerous papers using recurrent neural networks for language modeling \cite{bengio2006neural,mikolov2010recurrent,mikolov2012context,ryan2014multimodal}. We build most directly on top of \cite{bengio2006neural,mikolov2010recurrent,mikolov2012context} that use RNNs to learn word context. Several models use other sources of contextual information to help inform the language model \cite{mikolov2012context,ryan2014multimodal}. Despite its success, RNNs still have difficulty capturing long-range relationships in sequential modeling~\cite{bengio1994learning}. One solution is Long Short-Term Memory (LSTM) networks~\cite{hochreiter1997long,sutskever2011generating,ryan2014multimodal}, which use ``gates'' to control gradient back-propagation explicitly and allow for the learning of long-term interactions. However, the main focus of this paper is to show that the hidden layers learned by ``translating'' between multiple modalities can already discover rich structures in the data and learn long distance relations in an automatic, data-driven manner.

%
%

\section{Approach}
\label{sec:approach}
In this section we describe our approach using recurrent neural networks. Our goals are twofold. First, we want to be able to generate sentences given a set of visual observations or features. Specifically, we want to compute the probability of a word $w_t$ being generated at time $t$ given the set of previously generated words $W_{t-1} = {w_1,\ldots,w_{t-1}}$ and the observed visual features $V$. Second, we want to enable the capability of computing the likelihood of the visual features $V$ given a set of spoken or read words $W_t$ for generating visual representations of the scene or for performing image search. To accomplish both of these tasks we introduce a set of latent variables $U_{t-1}$ that encodes the visual interpretation of the previously generated or read words $W_{t-1}$. As we demonstrate later, the latent variables $U$ play the critical role of acting as a long-term visual memory of the words that have been previously generated or read.

Using $U$, our goal is to compute $P(w_t |V, W_{t-1}, U_{t-1})$ and $P(V | W_{t-1}, U_{t-1})$. Combining these two likelihoods together our global objective is to maximize,
\begin{eqnarray}\label{eqn:global}
\lefteqn{P(w_t, V | W_{t-1}, U_{t-1} )} & & \nonumber \\
& &  = P(w_t |V, W_{t-1}, U_{t-1})P(V | W_{t-1}, U_{t-1}).
\end{eqnarray}
That is, we want to maximize the likelihood of the word $w_t$ and the observed visual features $V$ given the previous words and their visual interpretation. Note that in previous papers \cite{mikolov2012context,mao2014explain} the objective was only to compute $P(w_t |V, W_{t-1})$ and not $P(V | W_{t-1})$.

\subsection{Model structure}

Our recurrent neural network model structure builds on the prior models proposed by \cite{mikolov2010recurrent,mikolov2012context}. Mikolov \cite{mikolov2010recurrent} proposed a RNN language model shown by the green boxes in Figure \ref{fig:model}(a). The word at time $t$ is represented by a vector $\wf_t$ using a ``one hot'' representation. That is, $\wf_t$ is the same size as the word vocabulary with each entry having a value of 0 or 1 depending on whether the word was used. The output $\wtf_t$ contains the likelihood of generating each word. The recurrent hidden state $\ssf$ provides context based on the previous words. However, $\ssf$ typically only models short-range interactions due to the problem of vanishing gradients \cite{bengio1994learning,mikolov2010recurrent}. This simple, yet effective language model was shown to provide a useful continuous word embedding for a variety of applications \cite{mikolov2013efficient}.

Following \cite{mikolov2010recurrent}, Mikolov \etal \cite{mikolov2012context} added an input layer $\vf$ to the RNN shown by the white box in Figure \ref{fig:model}. This layer may represent a variety of information, such as topic models or parts of speech \cite{mikolov2012context}. In our application, $\vf$ represents the set of observed visual features. We assume the visual features $\vf$ are constant. These visual features help inform the selection of words. For instance, if a cat was detected, the word ``cat'' is more likely to be spoken. Note that unlike \cite{mikolov2012context}, it is not necessary to directly connect $\vf$ to $\wtf$, since $\vf$ is static for our application. In \cite{mikolov2012context} $\vf$ represented dynamic information such as parts of speech for which $\wtf$ needed direct access. We also found that only connecting $\vf$ to half of the $\ssf$ units provided better results, since it allowed different units to specialize on modeling either text or visual features.

The main contribution of this paper is the addition of the recurrent visual hidden layer $\uf$, blue boxes in Figure \ref{fig:model}(a). The recurrent layer $\uf$ attempts to reconstruct the visual features $\vf$ from the previous words, \ie $\vtf \approx \vf$. The visual hidden layer is also used by $\wtf_t$ to help in predicting the next word. That is, the network can compare its visual memory of what it has already said $\uf$ to what it currently observes $\vf$ to predict what to say next. At the beginning of the sentence, $\uf$ represents the prior probability of the visual features. As more words are observed, the visual feature likelihoods are updated to reflect the words' visual interpretation. For instance, if the word ``sink'' is generated, the visual feature corresponding to sink will increase. Other features that correspond to stove or refrigerator might increase as well, since they are highly correlated with sink.

\begin{figure}
  \centering
  \includegraphics[width=\linewidth]{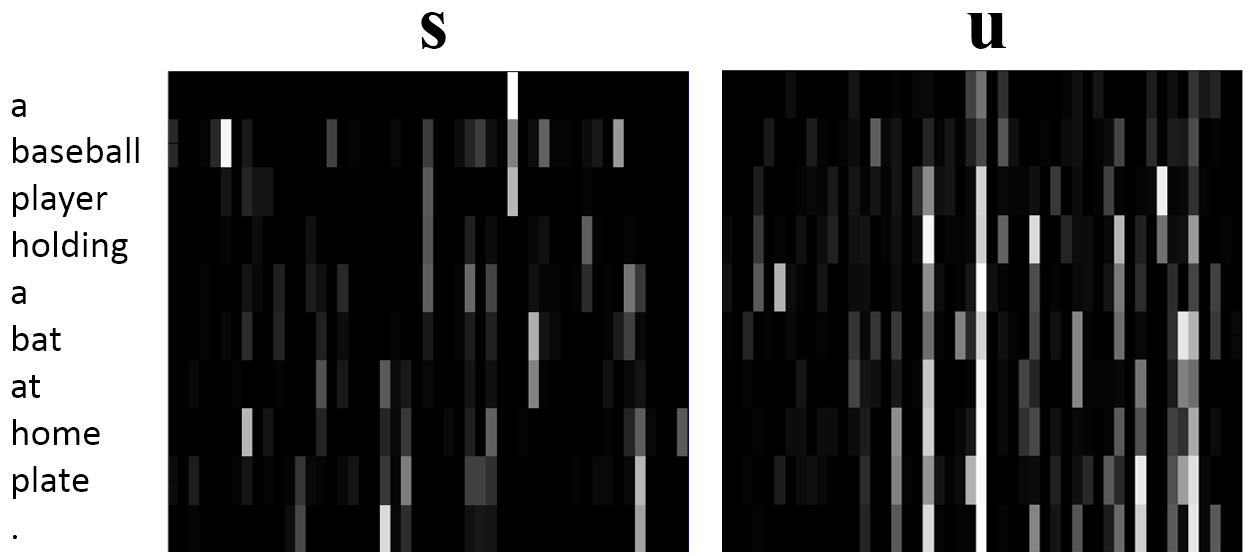}\\
  \caption{Illustration of the hidden units $\ssf$ and $\uf$ activations through time (vertical axis). Notice that the visual hidden units $\uf$ exhibit long-term memory through the temporal stability of some units, where the hidden units $\ssf$ change significantly each time step.}\label{fig:units}
\end{figure}

A critical property of the recurrent visual features $\uf$ is their ability to remember visual concepts over the long term. The property arises from the model structure. Intuitively, one may expect the visual features shouldn't be estimated until the sentence is finished. That is, $\uf$ should not be used to estimate $\vf$ until $\wf_t$ generates the end of sentence token. However, in our model we force $\uf$ to estimate $\vf$ at every time step to help in remembering visual concepts. For instance, if the word ``cat'' is generated, $\uf_t$ will increase the likelihood of the visual feature corresponding to cat. Assuming the ``cat'' visual feature in $\vf$ is active, the network will receive positive reinforcement to propagate $\uf$'s memory of ``cat'' from one time instance to the next. Figure \ref{fig:units} shows an illustrative example of the hidden units $\ssf$ and $\uf$. As can be observed, some visual hidden units $\uf$ exhibit longer temporal stability.

Note that the same network structure can predict visual features from sentences or generate sentences from visual features. For generating sentences (Fig. \ref{fig:model}(b)), $\vf$ is known and $\vtf$ may be ignored. For predicting visual features from sentences (Fig. \ref{fig:model}(c)), $\wf$ is known, and $\ssf$ and $\vf$ may be ignored. This property arises from the fact that the words units $\wf$ separate the model into two halves for predicting words or visual features respectively. Alternatively, if the hidden units $\ssf$ were connected directly to $\uf$, this property would be lost and the network would act as a normal auto-encoder \cite{vincent2008extracting}.

\subsection{Implementation details}

In this section we describe the details of our language model and how we learn our network.

\subsection{Language Model}

Our language model typically has between 3,000 and 20,000 words. While each word may be predicted independently, this approach is computationally expensive. Instead, we adopted the idea of word classing~\cite{mikolov2010recurrent} and factorized the distribution into a product of two terms:
\begin{equation}\label{eqn:global}
P(w_t |\cdot) = P(c_t |\cdot) \times P(w_t |c_t,\cdot).
\end{equation}
$P(w_t |\cdot)$ is the probability of the word, $P(c_t |\cdot)$ is the probability of the class. The class label of the word is computed in an unsupervised manner, grouping words of similar frequencies together. Generally, this approach greatly accelerates the learning process, with little loss of perplexity. The predicted word likelihoods are computed using the standard soft-max function. After each epoch, the perplexity is evaluated on a separate validation set and the learning reduced (cut in half in our experiments) if perplexity does not decrease.

In order to further reduce the perplexity, we combine the RNN model's output with the output from a Maximum Entropy language model~\cite{mikolov2011strategies}, simultaneously learned from the training corpus. For all experiments we fix how many words to look back when predicting the next word used by the Maximum Entropy model to three.

For any natural language processing task, pre-processing is crucial to the final performance. For all the sentences, we did the following two steps before feeding them into the RNN model. 1) Use Stanford CoreNLP Tool to tokenize the sentences. 2) Lower case all the letters. 

\subsection{Learning}

For learning we use the Backpropagation Through Time (BPTT) algorithm \cite{williams1989experimental}. Specifically, the network is unrolled for several words and standard backpropagation is applied. Note that we reset the model after an End-of-Sentence (EOS) is encountered, so that prediction does not cross sentence boundaries. As shown to be beneficial in ~\cite{mikolov2010recurrent}, we use online learning for the weights from the recurrent units to the output words. The weights for the rest of the network use a once per sentence batch update. The activations for all units are computed using the sigmoid function $\sigma(z) = 1 / (1 + \mathrm{exp}(-z))$ with clipping, except the word predictions that use soft-max. We found that Rectified Linear Units (ReLUs)~\cite{krizhevsky2012imagenet} with unbounded activations were numerically unstable and commonly ``blew up'' when used in recurrent networks.

We used the open source RNN code of~\cite{mikolov2010recurrent} and the Caffe framework~\cite{jia2014caffe} to implement our model. A big advantage of combining the two is that we can jointly learn the word and image representations: the error from predicting the words can be directly backpropagated to the image-level features. However, deep convolution neural networks require large amounts of data to train on, but the largest sentence-image dataset has only $\sim$80K images \cite{lin2014microsoft}. Therefore, instead of training from scratch, we choose to fine-tune from the pre-trained 1000-class ImageNet model \cite{deng2009imagenet} to avoid potential over-fitting. In all experiments, we used the 4096D 7th full-connected layer output as the visual input $\vf$ to our model.

\section{Results}
In this section we evaluate the effectiveness of our bi-directional RNN model on multiple tasks. We begin by describing the datasets used for training and testing, followed by our baselines. Our first set of evaluations measure our model's ability to generate novel descriptions of images. Since our model is bi-directional, we evaluate its performance on both the sentence retrieval and image retrieval tasks. For addition results please see the supplementary material.


\begin{table*}[t]
\centering
\small
\begin{tabular}{c|c|c|c|c|c|c|c|c|c|c|c|c}
    \hline
     & \multicolumn{3}{c|}{PASCAL} & \multicolumn{3}{c|}{Flickr 8K} & \multicolumn{3}{c|}{Flickr 30K} & \multicolumn{3}{c}{MS COCO} \\
    \cline{2-13}
    & PPL & BLEU & METR & PPL & BLEU & METR & PPL & BLEU & METR & PPL & BLEU & METR \\
    \hline
    Midge~\cite{mitchell2012midge} & - & 2.89 & 8.80 & \multicolumn{9}{c}{-} \\
    Baby Talk~\cite{kulkarni2011baby} & - & 0.49 & 9.69 & \multicolumn{9}{c}{-} \\
    \hline
    RNN & 36.79 & 2.79 & 10.08 & 21.88 & 4.86 & 11.81 & 26.94 & 6.29 & 12.34 & 18.96 & 4.63 & 11.47 \\
    RNN+IF & 30.04 & 10.16 & 16.43 & 20.43 & 12.04 & 17.10 & 23.74 & 10.59 & 15.56 & 15.39 & 16.60 & 19.24 \\
    RNN+IF+FT & 29.43 & 10.18 & 16.45 & - & - & - & - & - & - & 14.90 & 16.77 & 19.41 \\
    \hline
    Our Approach & 27.97 & 10.48 & 16.69 & 19.24 & 14.10 & 17.97 & 22.51 & 12.60 & 16.42 & 14.23 & 18.35 &  20.04 \\
    Our Approach + FT & 26.95 & 10.77 & 16.87 & - & - & - & - & - & - & 13.98 & 18.99 & 20.42 \\
    \hline
    Human & - & 22.07 & 25.80 & - & 22.51 & 26.31 & - & 19.62 & 23.76 & - & 20.19 & 24.94 \\
    \hline
\end{tabular}

\caption{Results for novel sentence generation for PASCAL 1K, Flickr 8K, FLickr 30K and MS COCO. Results are measured using perplexity (PPL), BLEU (\%) \cite{papineni2002bleu} and METEOR (METR, \%) \cite{banerjee2005meteor}. When available results for Midge~\cite{mitchell2012midge} and BabyTalk~\cite{kulkarni2011baby} are provided. Human agreement scores are shown in the last row. See the text for more details. \label{tab:sentgen}}
\end{table*}

\subsection{Datasets}

For evaluation we perform experiments on several standard datasets that are used for sentence generation and the sentence-image retrieval task:

\paragraph{PASCAL 1K~\cite{Rashtchian2010}} The dataset contains a subset of images from the PASCAL VOC challenge. For each of the 20 categories, it has a random sample of 50 images with 5 descriptions provided by Amazon's Mechanical Turk (AMT).

\paragraph{Flickr 8K and 30K~\cite{Rashtchian2010}} These datasets consists of 8,000 and 31,783 images collected from Flickr respectively. Most of the images depict humans participating in various activities. Each image is also paired with 5 sentences. These datasets have a standard training, validation, and testing splits.

\paragraph{MS COCO~\cite{lin2014microsoft}} The Microsoft COCO dataset contains 82,783 training images
and 40,504 validation images each with $\sim$5 human generated descriptions. The images are collected from Flickr by searching for common object categories, and typically contain multiple objects with significant contextual information. We downloaded the version which contains $\sim$40K annotated training images and $\sim$10K validation images for our experiments.


\begin{table*}[t]
\centering
\small
\begin{tabular}{c|c|c|c|c|c|c|c|c}
    \hline
     & \multicolumn{4}{c|}{Sentence Retrieval} & \multicolumn{4}{c}{Image Retrieval} \\
    \cline{2-9}
    & R@1 & R@5 & R@10 & Mean $r$ & R@1 & R@5 & R@10 & Mean $r$  \\
    Random Ranking & 4.0 & 9.0 & 12.0 & 71.0 & 1.6 & 5.2 & 10.6 & 50.0 \\
    \hline
    DeViSE \cite{frome2013devise} & 17.0 & 57.0 & 68.0 & 11.9 & 21.6 & 54.6 & 72.4 & 9.5 \\
    SDT-RNN \cite{socher2013grounded} & 25.0 & 56.0 & 70.0 & 13.4 & 35.4 & 65.2 & 84.4 & 7.0 \\
    DeepFE \cite{karpathy2014deep} & 39.0 & 68.0 & 79.0 & 10.5 & 23.6 & 65.2 & 79.8 & 7.6 \\
    \hline
    RNN+IF & 31.0 & 68.0 & 87.0 & 6.0 & 27.2 & 65.4 & 79.8 & 7.0 \\
    Our Approach (T) & 25.0 & 71.0 & 86.0 & 5.4 & 28.0 & 65.4 & 82.2 & 6.8 \\
    Our Approach (T+I) & 30.0 & 75.0 & 87.0 & 5.0 & 28.0 & 67.4 & 83.4 & 6.2 \\
    \hline
\end{tabular}

\caption{PASCAL image and sentence retrieval experiments. The protocol of \cite{socher2013grounded} was followed. Results are shown for recall at 1, 5, and 10 recalled items and the mean ranking (Mean r) of ground truth items.\label{tab:pcl}}
\end{table*}

\begin{table*}[t]
\centering
\small
\begin{tabular}{c|c|c|c|c|c|c|c|c}
    \hline
     & \multicolumn{4}{c|}{Sentence Retrieval} & \multicolumn{4}{c}{Image Retrieval} \\
    \cline{2-9}
    & R@1 & R@5 & R@10 & Med $r$ & R@1 & R@5 & R@10 & Med $r$  \\
    Random Ranking & 0.1 & 0.6 & 1.1 & 631 & 0.1 & 0.5 & 1.0 & 500 \\
    \hline
    \cite{socher2013grounded} & 4.5 & 18.0 & 28.6 & 32 & 6.1 & 18.5 & 29.0 & 29 \\
    DeViSE \cite{frome2013devise} & 4.8 & 16.5 & 27.3 & 28 & 5.9 & 20.1 & 29.6 & 29 \\
    DeepFE \cite{karpathy2014deep} & 12.6 & 32.9 & 44.0 & 14 & 9.7 & 29.6 & 42.5 & 15 \\
    DeepFE+DECAF \cite{karpathy2014deep} & 5.9 & 19.2 & 27.3 & 34 & 5.2 & 17.6 & 26.5 & 32 \\
    RNN+IF & 7.2 & 18.7 & 28.7 & 30.5 & 4.5 & 15.34 & 24.0 & 39 \\
    Our Approach (T) & 7.6 & 21.1 & 31.8 & 27 & 5.0 & 17.6 & 27.4 & 33 \\
    Our Approach (T+I) & 7.7 & 21.0 & 31.7 & 26.6 & 5.2 & 17.5 & 27.9 & 31 \\
    \hline
	\cite{hodosh2013framing} & 8.3 & 21.6 & 30.3 & 34 & 7.6 & 20.7 & 30.1 & 38 \\
    RNN+IF & 5.5 & 17.0 & 27.2 & 28 & 5.0 & 15.0 & 23.9 & 39.5 \\
    Our Approach (T) & 6.0 & 19.4 & 31.1 & 26 & 5.3 & 17.5 & 28.5 & 33 \\
    Our Approach (T+I) & 6.2 & 19.3 & 32.1 & 24 & 5.7 & 18.1 & 28.4 & 31 \\
    \hline
    M-RNN \cite{mao2014explain} & 14.5 & 37.2 & 48.5 & 11 & 11.5 & 31.0 & 42.4 & 15 \\
    RNN+IF & 10.4 & 30.9 & 44.2 & 14 & 10.2 & 28.0 & 40.6 & 16 \\
    Our Approach (T) & 11.6 & 33.8 & 47.3 & 11.5 & 11.4 & 31.8 & 45.8 & 12.5 \\
    Our Approach (T+I) & 11.7 & 34.8 & 48.6 & 11.2 & 11.4 & 32.0 & 46.2 & 11 \\
    \hline
\end{tabular}

\caption{Flickr 8K Retrieval Experiments. The protocols of \cite{socher2013grounded}, \cite{hodosh2013framing} and \cite{mao2014explain} are used respectively in each row. See text for details.\label{tab:f8k}}
\end{table*}

\begin{table*}[t]
\centering
\small
\begin{tabular}{c|c|c|c|c|c|c|c|c}
    \hline
     & \multicolumn{4}{c|}{Sentence Retrieval} & \multicolumn{4}{c}{Image Retrieval} \\
    \cline{2-9}
    & R@1 & R@5 & R@10 & Med $r$ & R@1 & R@5 & R@10 & Med $r$  \\
    Random Ranking & 0.1 & 0.6 & 1.1 & 631 & 0.1 & 0.5 & 1.0 & 500 \\
    \hline
    DeViSE \cite{frome2013devise} & 4.5 & 18.1 & 29.2 & 26 & 6.7 & 21.9 & 32.7 & 25 \\
    DeepFE+FT \cite{karpathy2014deep} & 16.4 & 40.2 & 54.7 & 8 & 10.3 & 31.4 & 44.5 & 13  \\
    RNN+IF & 8.0 & 19.4 & 27.6 & 37 & 5.1 & 14.8 & 22.8 & 47 \\
    Our Approach (T) & 9.3 & 23.8 & 24.0 & 28 & 6.0 & 17.7 & 27.0 & 35 \\
    Our Approach (T+I) & 9.6 & 24.0 & 27.2 & 25 & 7.1 & 17.9 & 29.0 & 31 \\
    \hline
    M-RNN \cite{mao2014explain} & 18.4 & 40.2 & 50.9 & 10 & 12.6 & 31.2 & 41.5 & 16 \\
    RNN+IF & 9.5 & 29.3 & 42.4 & 15 & 9.2 & 27.1 & 36.6 & 21 \\
    Our Approach (T) & 11.9 & 25.0 & 47.7 & 12 & 12.8 & 32.9 & 44.5 & 13 \\
    Our Approach (T+I) & 12.1 & 27.8 & 47.8 & 11 & 12.7 & 33.1 & 44.9 & 12.5 \\
    \hline
\end{tabular}

\caption{Flickr 30K Retrieval Experiments.  The protocols of \cite{frome2013devise} and \cite{mao2014explain} are used respectively in each row. See text for details.}\label{tab:f30k}
\end{table*}

\subsection{RNN Baselines}
To gain insight into the various components of our model, we compared our final model with three RNN baselines. For fair comparison, the random seed initialization was fixed for all experiments. The the hidden layers $\ssf$ and $\uf$ sizes are fixed to 100. We tried increasing the number of hidden units, but results did not improve. For small datasets, more units can lead to overfitting.

\paragraph{RNN based Language Model (RNN)} This is the basic RNN language model developed by ~\cite{mikolov2010recurrent}, which has no input visual features.
\paragraph{RNN with Image Features (RNN+IF)} This is an RNN model with image features feeding into the hidden layer inspired by \cite{mikolov2012context}. As described in Section \ref{sec:approach} $\vf$ is only connected to $\ssf$ and not $\wtf$. For the visual features $\vf$ we used the 4096D 7th Layer output of BVLC reference Net~\cite{jia2014caffe} after ReLUs. This network is trained on the ImageNet 1000-way classification task \cite{deng2009imagenet}. We experimented with other layers (5th and 6th) but they do not perform as well. 

\paragraph{RNN with Image Features Fine-Tuned (RNN+FT)} This model has the same architecture as RNN+IF, but the error is back-propagated to the Convolution Neural Network~\cite{girshickCVPR14}. The CNN is initialized with the weights from the BVLC reference net. The RNN is initialized with the the pre-trained RNN language model. That is, the only randomly initialized weights are the ones from visual features $\vf$ to hidden layers $\ssf$. If the RNN is not pre-trained we found the initial gradients to be too noisy for the CNN. If the weights from $\vf$ to hidden layers $\ssf$ are also pre-trained the search space becomes too limited.
    Our current implementation takes $\sim$5 seconds to learn a mini-batch of size 128 on a Tesla K40 GPU. It is also crucial to keep track of the validation error and avoid overfitting. We observed this fine-tuning strategy is particularly helpful for MS COCO, but does not give much performance gain on Flickr Datasets before it overfits. The Flickr datasets may not provide enough training data to avoid overfitting.

After fine-tuning, we fix the image features again and retrain our model on top of it.

\begin{figure*}
  \centering
  \includegraphics[width=1.0\linewidth]{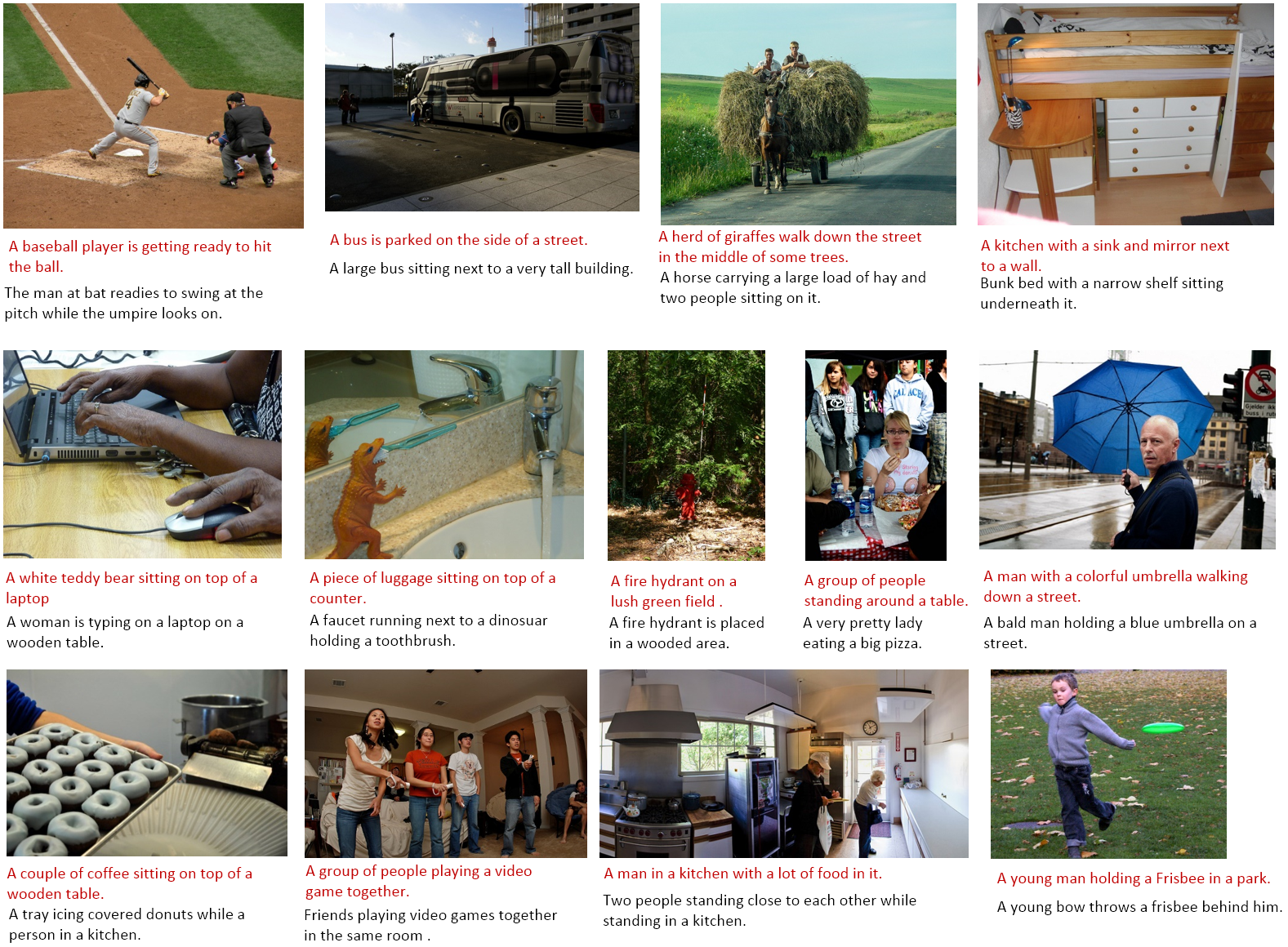}\\
  \caption{Qualitative results for sentence generation on the MS COCO dataset. Both a generated sentence (red) using (Our Approach + FT) and a human generated caption (black) are shown.}\label{fig:COCOqual}
\end{figure*}

\begin{figure*}
  \centering
  \includegraphics[width=0.83\linewidth]{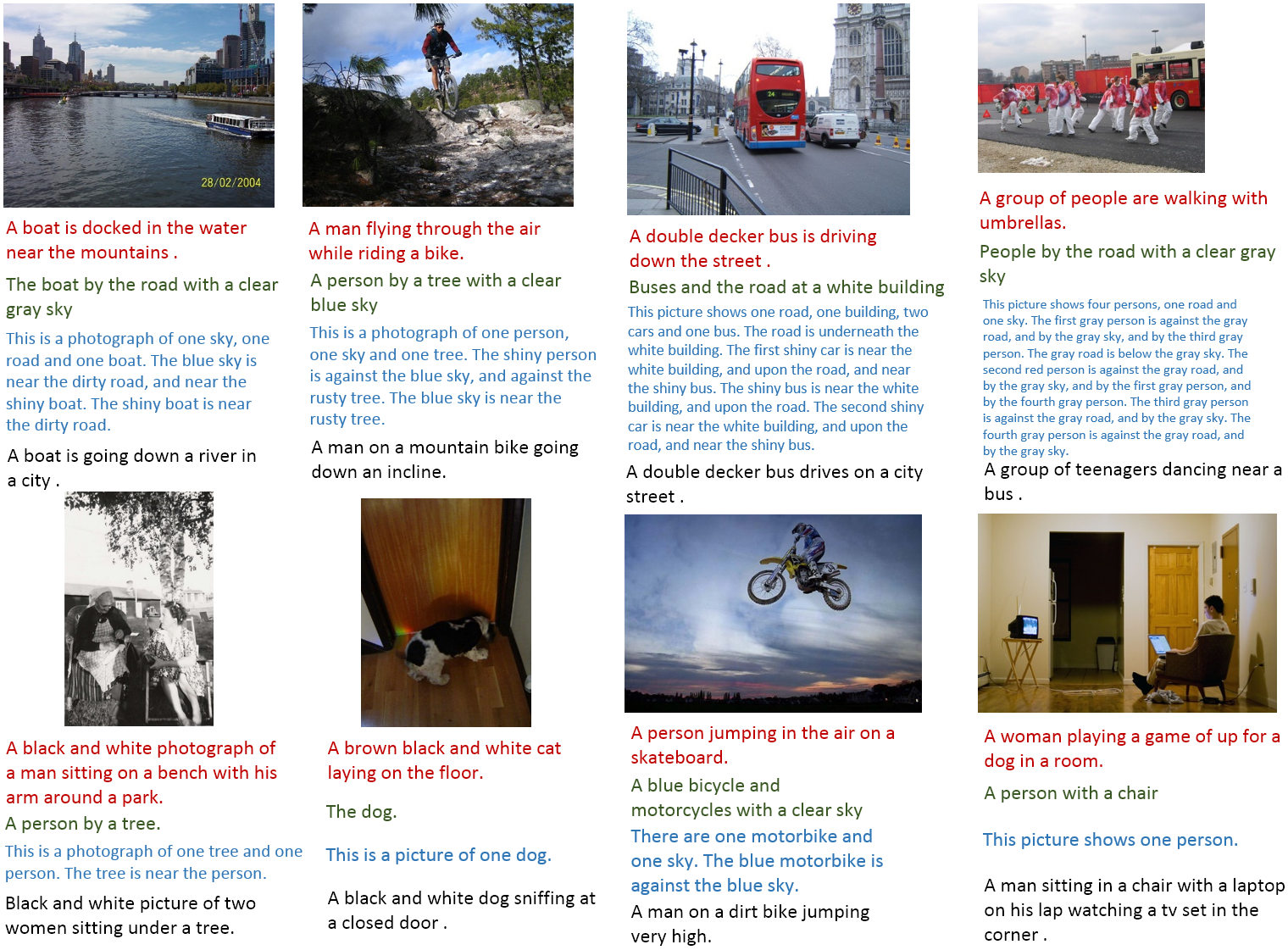}\\
  \caption{Qualitative results for sentence generation on the PASCAL 1K dataset. Generated sentences are shown for our approach (red), Midge \cite{mitchell2012midge} (green) and BabyTalk \cite{kulkarni2011baby} (blue). For reference, a human generated caption is shown in black.}\label{fig:PASCALqual}
\end{figure*}

\subsection{Sentence generation}

Our first set of experiments evaluate our model's ability to generate novel sentence descriptions of images. We experiment on all the image-sentence datasets described previously and compare to the RNN baselines and other previous papers~\cite{mitchell2012midge,kulkarni2011baby}. Since PASCAL 1K has a limited amount of training data, we report results trained on MS COCO and tested on PASCAL 1K. We use the standard train-test splits for the Flickr 8K and 30K datasets. For MS COCO we train and validate on the training set ($\sim$37K/$\sim$3K), and test on the validation set, since the testing set is not available. To generate a sentence, we first sample a target sentence length from the multinomial distribution of lengths learned from the training data, then for this fixed length we sample 100 random sentences, and use the one with the lowest loss (negative likelihood, and in case of our model, also reconstruction error) as output.

We choose three automatic metrics for evaluating the quality of the generated sentences, perplexity, BLEU \cite{papineni2002bleu} and METEOR \cite{banerjee2005meteor}. Perplexity measures the likelihood of generating the testing sentence based on the number of bits it would take to encode it. The lower the value the better. BLEU and METEOR were originally designed for automatic machine translation where they rate the quality of a translated sentences given several references sentences. We can treat the sentence generation task as the ``translation'' of images to sentences. For BLEU, we took the geometric mean of the scores from 1-gram to 4-gram, and used the ground truth length closest to the generated sentence to penalize brevity. For METEOR, we used the latest version\footnote{\url{http://www.cs.cmu.edu/~alavie/METEOR/}} (v1.5). For both BLEU and METEOR higher scores are better. For reference, we also report the consistency between human annotators (using 1 sentence as query and the rest as references)\footnote{We used 5 sentences as references for system evaluation, but leave out 4 sentences for human consistency. It is a bit unfair but the difference is usually around 0.01$\sim$0.02.}.

Results are shown in Table \ref{tab:sentgen}. Our approach significantly improves over both Midge \cite{mitchell2012midge} and BabyTalk \cite{kulkarni2011baby} on the PASCAL 1K dataset as measured by BLEU and METEOR. Several qualitative results for the three algorithms are shown in Figure \ref{fig:PASCALqual}. Our approach generally provides more naturally descriptive sentences, such as mentioning an image is black and white, or a bus is a ``double decker''. Midge's descriptions are often shorter with less detail and BabyTalk provides long, but often redundant descriptions. Results on Flickr 8K and Flickr 30K are also provided.

On the MS COCO dataset that contains more images of high complexity we provide perplexity, BLEU and METEOR scores. Surprisingly our BLEU and METEOR scores (18.99 \& 20.42) are just slightly lower than the human scores (20.19 \& 24.94). The use of image features (RNN + IF) significantly improves performance over using just an RNN language model. Fine-tuning (FT) and our full approach provide additional improvements for all datasets. For future reference, our final model gives BLEU-1 to BLEU-4 (with penalty) as 60.4\%, 26.4\%, 12.6\% and 6.5\%, compared to human consistency 65.9\%, 30.5\%, 13.6\% and 6.0\%. Qualitative results for the MS COCO dataset are shown in Figure \ref{fig:COCOqual}. Note that since our model is trained on MS COCO, the generated sentences are generally better on MS COCO than PASCAL 1K.

It is known that automatic measures are only roughly correlated with human judgment \cite{elliot2014}, so it is also important to evaluate the generated sentences using human studies. We evaluated 1000 generated sentences on MS COCO by asking human subjects to judge whether it had better, worse or same quality to a human generated ground truth caption. 5 subjects were asked to rate each image, and the majority vote was recorded. In the case of a tie (2-2-1) the two winners each got half of a vote. We find $12.6\%$ and $19.8\%$ prefer our automatically generated captions to the human captions without (Our Approach) and with fine-tuning (Our Approach + FT) respectively. Less than $1\%$ of the subjects rated the captions as the same. This is an impressive result given we only used image-level visual features for the complex images in MS COCO.

\subsection{Bi-Directional Retrieval}
Our RNN model is bi-directional. That is, it can generate image features from sentences and sentences from image features. To evaluate its ability to do both, we measure its performance on two retrieval tasks. We retrieve images given a sentence description, and we retrieve a description given an image. Since most previous methods are capable of only the retrieval task, this also helps provide experimental comparison.

Following other methods, we adopted two protocols for using multiple image descriptions. The first one is to treat each of the $\sim$5 sentences individually. In this scenario, the rank of the retrieved ground truth sentences are used for evaluation. In the second case, we treat all the sentences as a single annotation, and concatenate them together for retrieval.

For each retrieval task we have two methods for ranking. First, we may rank based on the likelihood of the sentence given the image (T). Since shorter sentences naturally have higher probability of being generated, we followed~\cite{mao2014explain} and normalized the probability by dividing it with the total probability summed over the entire retrieval set. Second, we could rank based on the reconstruction error between the image's visual features $\vf$ and their reconstructed visual features $\vtf$ (I). Due to better performance, we use the average reconstruction error over all time steps rather than just the error at the end of the sentence. In Tables \ref{tab:f8k}, we report retrieval results on using the text likelihood term only (I) and its combination with the visual feature reconstruction error (T+I).

The same evaluation metrics were adopted from previous papers for both the tasks of sentence retrieval and image retrieval. They used R@K (K = 1, 5, 10) as the measurements, which are the recall rates of the (first) ground truth sentences (sentence retrieval task) or images (image retrieval task). Higher R@K corresponds to better retrieval performance. We also report the median/mean rank of the (first) retrieved ground truth sentences or images (Med/Mean r). Lower Med/Mean $r$ implies better performance. For Flickr 8K and 30K several different evaluation methodologies have been proposed. We report three scores for Flickr 8K corresponding to the methodologies proposed by \cite{socher2013grounded}, \cite{hodosh2013framing} and \cite{mao2014explain} respectively, and for Flickr 30K \cite{frome2013devise} and \cite{mao2014explain}.

Measured by Mean r, we achieve state-of-the-art results on PASCAL 1K image and sentence retrieval (Table \ref{tab:pcl}). As shown in Tables \ref{tab:f8k} and \ref{tab:f30k}, for Flickr 8K and 30K our approach achieves comparable or better results than all methods except for the recently proposed DeepFE \cite{karpathy2014deep}. However, DeepFE uses a different set of features based on smaller image regions. If the same features are used (DeepFE+DECAF) as our approach, we achieve better results. We believe these contributions are complementary, and by using better features our approach may also show further improvement. In general ranking based on text and visual features (T + I) outperforms just using text (T). Please see the supplementary material for retrieval results on MS COCO.

\section{Discussion}

Image captions describe both the objects in the image and their relationships. An area of future work is to examine the sequential exploration of an image and how it relates to image descriptions. Many words correspond to spatial relations that our current model has difficultly in detecting. As demonstrated by the recent paper of \cite{karpathy2014deep} better feature localization in the image can greatly improve the performance of retrieval tasks and similar improvement might be seen in the description generation task.

In conclusion, we describe the first bi-directional model capable of the generating both novel image descriptions and visual features. Unlike many previous approaches using RNNs, our model is capable of learning long-term interactions. This arises from using a recurrent visual memory that learns to reconstruct the visual features as new words are read or generated. We demonstrate state-of-the-art results on the task of sentence generation, image retrieval and sentence retrieval on numerous datasets.

\section{Acknowledgements}

We thank Hao Fang, Saurabh Gupta, Meg Mitchell, Xiaodong He, Geoff Zweig, John Platt and Piotr Dollar for their thoughtful and insightful discussions in the creation of this paper.

{\small
\bibliographystyle{ieee}
\bibliography{RNN}
}

\end{document}